\documentclass{article}

\usepackage{spconf,amsmath,epsfig}
\usepackage{amsfonts}
\usepackage{amssymb}  
\usepackage[ruled,linesnumbered]{algorithm2e} 

\usepackage{float}
\usepackage{multirow}
\usepackage{booktabs}
\usepackage{url}

\usepackage[square,sort&compress,numbers]{natbib}

\title{Pixel-Level Change Detection Pseudo-Label Learning for Remote Sensing Change Captioning}
\name{Chenyang Liu\textsuperscript{1}, Keyan Chen\textsuperscript{1}, Zipeng Qi\textsuperscript{1}, Zili Liu\textsuperscript{1}, Haotian Zhang\textsuperscript{1}, Zhengxia Zou\textsuperscript{2}, Zhenwei Shi\textsuperscript{1,*}
\thanks{The work was supported by the National Key Research and Development Program of China (Grant No. 2022ZD0160401), the National Natural Science Foundation of China under Grants 62125102 and 623B2013, the Beijing Natural Science Foundation under Grant JL23005, and the Fundamental Research Funds for the Central Universities. *Corresponding author: Zhenwei Shi (e-mail: shizhenwei@buaa.edu.cn).}
}
\address{\textsuperscript{1}Image Processing Center, School of Astronautics, Beihang University\\
\textsuperscript{2} Department of Guidance, Navigation and Control, School of Astronautics, Beihang University
}
\begin{document}
%\ninept
%
\maketitle
\begin{abstract}
The existing Remote Sensing Image Change Captioning (RSICC) methods perform well in simple scenes but exhibit poorer performance in complex scenes. This limitation is primarily attributed to the model's constrained visual ability to distinguish and locate changes. Acknowledging the inherent correlation between change detection (CD) and RSICC tasks, we believe pixel-level CD is significant for describing the differences between images through language. Regrettably, the current RSICC dataset lacks readily available pixel-level CD labels. To address this deficiency, we leverage a model trained on existing CD datasets to derive CD pseudo-labels. We propose an innovative network with an auxiliary CD branch, supervised by pseudo-labels. Furthermore, a semantic fusion augment (SFA) module is proposed to fuse the feature information extracted by the CD branch, thereby facilitating the nuanced description of changes. Experiments demonstrate that our method achieves state-of-the-art performance and validate that learning pixel-level CD pseudo-labels significantly contributes to change captioning. 

% Our code will be available at: \emph{{https://github.com/Chen-Yang-Liu/Pix4Cap}}

% To address this deficiency, we leverage a model trained on extant CD datasets to derive CD pseudo-labels. We introduce an innovative network featuring an auxiliary CD branch, supervised by pseudo-labels. Furthermore, we propose a Semantic Fusion Augment (SFA) module to amalgamate feature information extracted by the CD branch, thereby facilitating a nuanced description of changes. Experimental results showcase the state-of-the-art performance of our proposed method, affirming that the acquisition of pixel-level CD pseudo-labels significantly contributes to change captioning.

\end{abstract}
\section{Introduction}
\label{sec:intro}
Remote sensing image (RSI) change interpretation technology \cite{Peng2020,zhang2024bifa}, such as change detection and change captioning, has played a pivotal role in providing valuable insights into environmental changes, urban development, and disaster monitoring \cite{chen2021building,chen2024rsprompter,qi2023implicit}. The RSI change captioning (CC) aiming to generate linguistic information describing changes has emerged as a promising avenue for enhancing the interpretability and communicative capacity of RSI. 

The existing RSICC methods predominantly adopt the encoder-decoder architecture \cite{NWPU_Captions,RSICap,Liu_2022,liu2024change_agent}. Pioneering efforts by Chouaf and Hoxha \textit{et al.} \cite{RSICC_1, RSICC_2} delved into the RSICC task, employing CNN-based visual encoders for bi-temporal image feature extraction and RNN-based decoders for the sequential generation of words. Extending the RSICC landscape, Liu \textit{et al.} \cite{RSICCformer} proposed a large LEVIR-CC dataset and the RSICCformer model, leveraging a dual-branch Transformer to enhance bi-temporal features and highlight the changes of interest. Addressing the multi-scale challenge, PSNet \cite{PSNet} extracts multi-scale features by combining progressive difference perception layers and scale-aware enhancement modules. Chang \textit{et al.} \cite{chg2cap} proposed an attentive encoder comprising hierarchical self-attention blocks and a residual block to extract visual embeddings. PromptCC \cite{PromptCC} efficiently utilizes a large language model as a language generator through a multi-prompt learning strategy determined by a coarse change classifier.
% Liu \textit{et al.} \cite{PromptCC} proposed

\begin{figure*}[ht]
	\centering
	\includegraphics[width=0.9\linewidth]{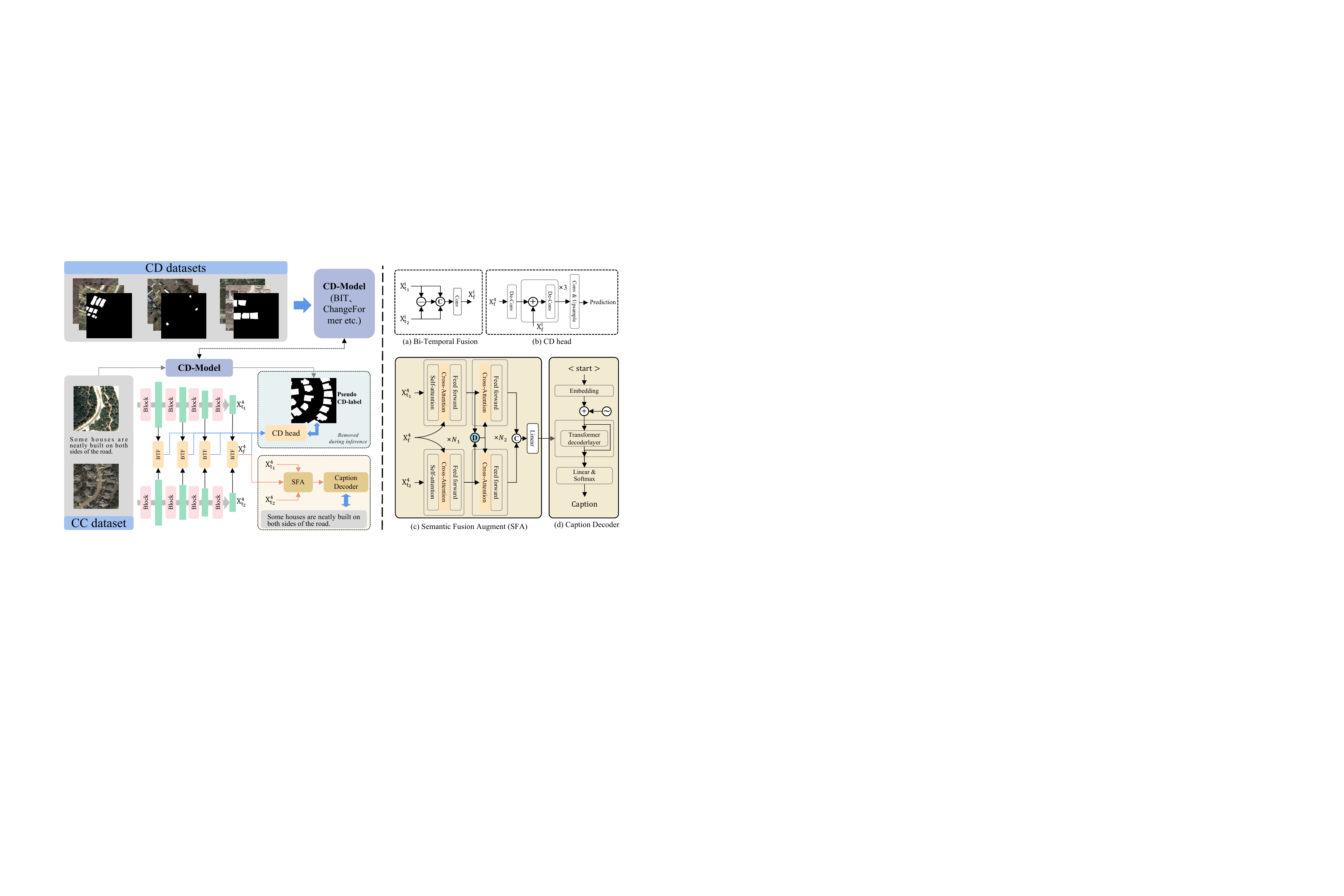}
    	 \caption{Illustration of our Pix4cap method. 
      \textbf{Left:} the overview of our model. The model comprises two branches: change captioning and auxiliary change detection. The SFA module acts as a pivotal bridge between the two branches. \textbf{Right:} the structure of some important modules.
      }
        \label{fig:overall}
    \vspace{-10pt}
\end{figure*}

Despite significant advancements, current RSICC methods face challenges in achieving robust performance, especially when confronted with scenarios characterized by subtle alterations in building structures, adverse lighting conditions, and low-contrast images. The degradation in performance can be largely attributed to the model's constrained visual capabilities, which hampers its ability to identify relevant changes while disregarding irrelevant ones accurately. Given the persistence of these challenges, it becomes imperative to explore innovative strategies that can improve the performance of RSICC models.

Acknowledging the intrinsic connection between change detection (CD) and RSICC, we propose that integrating pixel-level CD outcomes holds substantial promise. Pixel-level CD entails the meticulous identification and precise localization of changes at the pixel level, offering invaluable insights for generating nuanced and precise descriptions of differences between images. However, a significant impediment to realizing this potential lies in the absence of pixel-level CD labels within the current RSICC dataset.

In this paper, to bridge this crucial gap, we propose to leverage a model pre-trained on existing CD datasets to derive Change masks as our CD pseudo-labels. Then we propose a novel RSICC network with an auxiliary CD branch supervised by pseudo-labels. This pseudo-labeling strategy serves as a stepping stone toward enhancing the model's capability to perceive changes accurately, particularly in scenes where previous RSICC methods encounter difficulties. Furthermore, we present a Semantic Fusion Augment (SFA) module to seamlessly integrate feature information from the CD branch into the caption branch. Experiments demonstrate that our proposed method achieves state-of-the-art performance, underscoring the efficacy of CD pseudo-label learning as a viable enhancement strategy for the RSICC task.
% To this end, 

% In the subsequent sections, we elaborate on our proposed methodology, present experimental results demonstrating its efficacy, and discuss the implications and potential avenues for further research in the domain of remote sensing change captioning.
% weaknesses
% To further improve the model performance, one key is to improve the model's ability to identify visual changes of interest in complex scenes. The RS images contain objects of different scales \cite{Liu_2022}, which poses significant challenges in identifying and describing object attributes and complex relationships of changed objects. However, previous methods still have weaknesses in sufficiently extracting and utilizing multi-scale information in bi-temporal RS images. 

\section{Methodology}
\label{sec:method}
\subsection{Overview}
Fig. \ref{fig:overall} illustrates the overall structure of our proposed Pix4Cap (Pixel change detection for change Captioning). Pix4Cap comprises two branches dedicated to CD and RSICC, respectively. In the proposed Pix4Cap model, the SFA module acts as a pivotal bridge, facilitating the integration of features extracted by the auxiliary CD branch into the RSICC branch and contributing to the generation of change captions. Besides, the auxiliary CD branch is supervised by the pseudo-labels generated by the pre-trained CD models, such as BIT \cite{BIT} and ChangeFormer \cite{bandara2022transformer}.
% contribute to the generation of 

\subsection{Change Detection Branch}
The architecture of the CD branch mirrors that of UNet. Bi-temporal images serve as inputs to the backbone network, yielding four-scale bi-temporal features $X_{t_1}^i$, $X_{t_2}^i$ (i=1,2,3,4). Subsequently, the bi-temporal fusion (BTF) module processes these features alongside their differential features to highlight changes of interest. The CD head then predicts the change map by progressively decoding multi-level features from the BTF module. The structure of the BTF module and CD head are illustrated in Fig. \ref{fig:overall} (a) and (b). For the multi-level bi-temporal features $X^i_{t_1}$ and $X^i_{t_2}$ extracted by the backbone, the processing of the BTF module can be expressed as:
\begin{align}
\mathrm{\Phi_{BTF}}(X^i_{t_1}, X^i_{t_2}) &= {\rm Conv}({\rm Concat}([X^i_{t_1}, X^i_{t_2},X^i_{d}]))\\
X^i_{d} &= X^i_{t_{2}} - X^i_{t_{1}}
\end{align}
In the CD head, through deconvolution, the multi-level features $X^i_{f}$ are progressively integrated from bottom to top, generating accurate change masks. 

\vspace{-8pt}
\subsection{Change Caption Branch}
We anticipate that the CD branch will enhance the visual perception capability of the backbone, and the features from the BTF module serving as the neck of the CD branch can provide valuable supplementary information for the change captioning. To this end, we propose an SFA module based on the multi-head cross attention mechanism to seamlessly integrate the feature $X_f^4$ from the last BTF module into the RSICC branch, as illustrated in Fig. \ref{fig:overall} (c), where $D$ represents the following process:
\begin{align}
X_{dif} = W(X_{t_{2}} - X_{t_{1}}) +b+ {\rm Cos}(X_{t_{1}}, X_{t_{2}})
\end{align}
where $W$ and $b$ denote the projection matrix and bias of a linear layer, and $\rm Cos$ represents the operation of calculating cosine similarity. The subtraction operation and cosine similarity operation are used to measure the similarity between bi-temporal features from different perspectives. We believe this helps locate changed areas and capture distinctive features revealing changes of interest. For the input query $X_q$ and $X_c$ that needs to be integrated, the multi-head cross attention can be formulated as follows:
\begin{align}
\mathrm{\Phi_{MCA}}(Q,K,V) = {\rm Concat}(h_1;...;h_n)W^o \\
h_i = {\rm Att}(X_q W^Q_i, X_c W^K_i, X_c W^V_j)\\
\mathrm{\Phi_{Att}}(Q_i, K_i, V_i) = {\rm Softmax}({\frac {{Q_i}{K_i}^T} {\sqrt{d}}})V_i
\end{align}
where $W^Q_i \in R^{C \times d}, W^K_i \in R^{C \times d}, W^V_i \in R^{C \times d}$, and $W^o \in R^{hd \times C}$ are learnable parameter matrices, $d$ is the scaling factor, $n$ is the number of heads.

% Following the extraction of visual embeddings by the SFA module, 
Following the SFA module, we employ a Transformer decoder with a residual connection as the caption generator. This decoder utilizes visual embeddings to generate accurate and detailed sentences describing changes between images.

\begin{table*}[!htb]%[ht] \small%[htbp]%[!t]  %\small
\renewcommand{\arraystretch}{1.0}
\caption{Comparisons experiments on the LEVIR-CC dataset, where the bolded results are the best.}
\label{tab:Comparisons_other_methods}
\centering
% \resizebox{0.9\linewidth}{32mm}{
\begin{tabular}{c|c c c c c c c| c}
	\toprule%[1pt]
	%\hline
	Method & BLEU-1 & BLEU-2 & BLEU-3 & BLEU-4 & METEOR & ROUGE$_L$ & CIDEr & $S^*_m$\\
	\midrule
	{Capt-Rep-Diff \cite{robust_CC}} & 72.90 & 61.98 & 53.62 & 47.41 & 34.47 & 65.64 & 110.57 & 64.52\\
	{Capt-Att \cite{robust_CC}} & 77.64 & 67.40 & 59.24 & 53.15 & 36.58 & 69.73 & 121.22 & 70.17\\
	{Capt-Dual-Att \cite{robust_CC}} & 79.51 & 70.57 & 63.23 & 57.46 & 36.56 & 70.69 & 124.42 & 72.28\\
	{DUDA \cite{robust_CC}} & 81.44 & 72.22 & 64.24 & 57.79 & 37.15 & 71.04 & 124.32 & 72.58\\
	{MCCFormer-S \cite{MCCformer}} & 79.90 & 70.26 & 62.68 & 56.68 & 36.17 & 69.46 & 120.39 & 70.68\\
	{MCCFormer-D \cite{MCCformer}} & 80.42 & 70.87 & 62.86 & 56.38 & 37.29 & 70.32 & 124.44 & 72.11\\
	{RSICCFormer \cite{RSICCformer}} & {84.72} & {76.27} & {68.87} & {62.77} & {39.61} & {74.12} & {134.12} & 77.65\\
     {PSNet \cite{PSNet}} & 83.86 & 75.13 & 67.89 & 62.11 & 38.80 & 73.60 & 132.62 & 76.78\\
	\midrule
 {Baseline} & 84.87 & 76.38 & 69.15 & 63.25 & 39.47 & 74.08 & 133.96 & 77.69\\

 {Pix4Cap} & \textbf{85.56} & \textbf{77.08} & \textbf{69.79} & \textbf{63.78} & \textbf{39.96} & \textbf{75.12} & \textbf{136.76} & \textbf{78.91}\\
 
	\bottomrule%[1pt]
\end{tabular}
% }
\end{table*}

\vspace{-8pt}
\section{Experiments}
\label{sec:experiment}
\subsection{Experimental Setup}
% \textbf{Dataset:} 
\textbf{Quantitative Results:} 
We conduct experiments on the LEVIR-CC dataset \cite{RSICCformer}, a large RSICC dataset with 10077 image pairs. Among these, 5038 image pairs are changed, while 5039 image pairs are unchanged. Each image pair of the dataset has five annotated captions, totaling 50,385 descriptive captions, describing diverse ground surface transformations like buildings, roads, and vegetation.
% The images cover 20 regions of Texas, USA.

\noindent \textbf{Loss Function}:
The model parameters are updated in a supervised learning framework during the training phase. The cross-entropy loss function is used to compute both CD loss and caption loss as follows:
\begin{equation}
\mathcal{L}_{{det}} = - \frac{1}{N} \sum_{n=1}^{N} \sum_{j=1}^{C} y_{n}^{(j)} \log(p_{n}^{(j)})
\end{equation}
where $C$ is the number of classes. $N$ are the number of the change mask pixels. $\boldsymbol{p}_{n}$ = [ $p_{n}^{(0)},...,p_{cls}^{(C)}$] is the predicted probability vector. $\boldsymbol{{y}}_{n}$ = [ ${y}_{n}^{(0)}, ..., {y}_{n}^{(C)}$] is the one-hot vector of the ground-truth.
\begin{equation}
\mathcal{L}_{cap} = - \sum_{t=1}^{L} \sum_{v=1}^{V} \widetilde{y}_t^{(v)} \log{p_t^{(v)}}
\end{equation}
where $\boldsymbol{p}_t$ = [ $p_t^{(1)},p_t^{(2)},...,p_t^{(V)}$] is the probability for predicting the $t$-th word. $\boldsymbol{{y}}_t$ = [ ${y}_t^{(1)}, {y}_t^{(2)},... {y}_t^{(V)}$] is the $t$-th ground-truth word embedding. $V$ is the vocabulary size, and $L$ is the caption length. 

Mathematically, the final total loss is as follows:
\begin{equation}
\mathcal{L}_{total} =  {\mathcal{L}_{det}} + {\mathcal{L}_{cap}}
\label{Equation:loss_total}
\end{equation}

\noindent \textbf{Training Details}:
The pre-trained BIT model \cite{BIT} is utilized to generate CD pseudo-labels for each image pair of the LEVIR-CC dataset. ResNet32 is employed as the backbone to extract visual features. Our model is implemented on PyTorch and trained using a single NVIDIA RTX 4090 GPU. During the training stage, we minimized the two losses and optimized the model parameters using the Adam optimizer. 

\noindent \textbf{Evaluation Metrics}:
Following previous change captioning works \cite{RSICC_1,RSICC_2,RSICCformer}, for the evaluation of generated captions, we use metrics including BLEU-N (N=1,2,3,4), ROUGE$_L$, METEOR, CIDEr, and $S^*_m$ \cite{Avg_metric} to quantitatively evaluate the quality of the generated sentences.

% \subsection{Experiment Results}
\vspace{-8pt}
\subsection{Comparison with the State-of-the-Art}
\label{ssec:comparison}
We conduct ablation experiments and report the quantitative comparison results between our method and previous RSICC methods, including four LSTM-based methods (Capt-Rep-Diff \cite{robust_CC}, Capt-Att \cite{robust_CC}, Capt-Dual-Att \cite{robust_CC}, DUDA \cite{robust_CC}) and four Transformer-based methods (MCCFormers-S \cite{MCCformer}, MCCFormers-D \cite{MCCformer}, RSICCFormer \cite{RSICCformer}, PSNet \cite{PSNet}). 

The experimental results are shown in Table \ref{tab:Comparisons_other_methods}. To confirm the effectiveness of CD pseudo-label learning, we set a baseline, which removes the CD branch and the first cross-attention operation of the SFA module. The ablation result verifies that CD pseudo-label learning can effectively improve the model performance on generating accurate captions, resulting in a $1.22\%$ improvement on the $S_m^*$ metric, establishing a new state-of-the-art (SOTA) performance. Additionally, we can see that our Pix4Cap model achieves SOTA performance and outperforms previous methods. Although the CD model only outputs pseudo-labels of changed buildings, this has significantly enhanced the model's performance. We believe that exploring the use of multi-class object CD pseudo-labels in the future will lead to even greater performance improvements in change captioning.

\vspace{-8pt}
\subsection{Quantitative Results}
\begin{figure}
	\centering
	\includegraphics[width=0.9\linewidth]{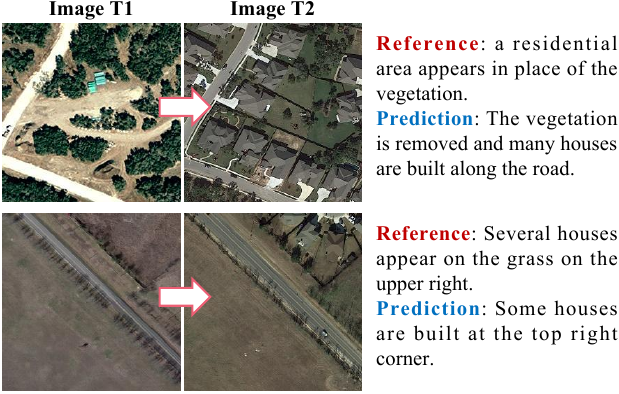}
	\caption{The comparison between one of five reference captions and the caption generated by our model.}
	\label{fig:cap}
 \vspace{-18pt}
\end{figure}

Fig. \ref{fig:cap} shows two examples of captions generated by our method. Despite illumination changes and low contrast between buildings and the surrounding environment, our method can accurately distinguish changes of interest and describe the changed object category, their spatial positions, and the relationship between the objects.

\vspace{-8pt}
\section{Conclusion}
% \vspace{-10pt}
\label{sec:conclusion}
This paper addresses the significant challenge of weak change recognition in current RSICC techniques, particularly in complex scenes. To overcome this obstacle, we leverage a pre-trained CD model to generate CD pseudo-labels for image pairs in the RSICC dataset. We propose the innovative Pix4Cap model, featuring an auxiliary CD branch supervised by pseudo-labels and a seamlessly integrated SFA module. Our approach ensures the effective integration of pixel-level CD information into the RSICC branch. The experiments validate the effectiveness of learning CD pseudo-labels in improving overall RSICC performance.

% \vfill
% \pagebreak

{

% \tiny 
\scriptsize
% \footnotesize
% \small
\bibliographystyle{IEEEbib}
\bibliography{refs}
}
\end{document}